\documentclass[letterpaper]{article}  
\usepackage{aaai2026}   
\usepackage{times}   
\usepackage{helvet}   
\usepackage{courier}   
\usepackage[hyphens]{url}   
\usepackage{graphicx}  
\usepackage{natbib}   
\usepackage{caption}  
\usepackage{algorithm}
\usepackage{algorithmic}

\usepackage{newfloat}
\usepackage{listings}
\DeclareCaptionStyle{ruled}{labelfont=normalfont,labelsep=colon,strut=off}  
\floatstyle{ruled}
\newfloat{listing}{tb}{lst}{}
\floatname{listing}{Listing}

\usepackage{adjustbox}
\usepackage{xcolor}
\usepackage{amssymb}
\usepackage{amsfonts}
\usepackage{xspace}
\usepackage{multirow}
\usepackage{amsmath}
\usepackage{booktabs}
 
\usepackage[inline]{enumitem}

\newcommand{\model}{KUnBR\xspace}
\newcommand{\fullmodel}{\textbf{K}nowledge Density-Guided \textbf{Un}learning via \textbf{B}locks \textbf{R}einsertion\xspace}

\newcommand{\aka}{\emph{a.k.a.,}\xspace}
\newcommand{\eg}{\emph{e.g.,}\xspace}

\newcommand{\ignore}[1]{}

\newcommand{\downarrowtext}{\ensuremath{\downarrow}}
\title{Beyond Superficial Forgetting: Thorough Unlearning \\ through Knowledge Density Estimation and Block Re-insertion}

\author{
    Feng Guo\textsuperscript{\rm 1}\equalcontrib,
    Yuntao Wen\textsuperscript{\rm 1}\equalcontrib,
    Shen Gao\textsuperscript{\rm 1}\thanks{Corresponding Author},
    Junshuo Zhang\textsuperscript{\rm 1},
    Shuo Shang\textsuperscript{\rm 1}
}
\affiliations{
     
    \textsuperscript{\rm 1}University of Electronic Science and Technology of China, Chengdu, China

    zxyz0051@163.com,
    shengao@uestc.edu.cn,
    \{yuntaowenx, hanblingz026, jedi.shang\}@gmail.com,
}

\usepackage{bibentry}

\begin{document}

\maketitle

\begin{abstract}
Machine unlearning, which selectively removes harmful knowledge from a pre-trained model without retraining from scratch, is crucial for addressing privacy, regulatory compliance, and ethical concerns in Large Language Models (LLMs). 
However, existing unlearning methods often struggle to thoroughly remove harmful knowledge, leaving residual harmful knowledge that can be easily recovered. 
To address these limitations, we propose \fullmodel (\model), a novel approach that first identifies layers with rich harmful knowledge and then thoroughly eliminates the harmful knowledge via re-insertion strategy. 
Our method introduces knowledge density estimation to quantify and locate layers containing the most harmful knowledge, enabling precise unlearning. 
Additionally, we design a layer re-insertion strategy that extracts and re-inserts harmful knowledge-rich layers into the original LLM, bypassing gradient obstruction caused by cover layers and ensuring effective gradient propagation during unlearning. 
Extensive experiments conducted on several unlearning and general capability benchmarks demonstrate that \model achieves state-of-the-art forgetting performance while maintaining model utility\footnote{Code available at \url{github.com/llmgfffffff/Beyond-Superficial-Forgetting-KUnBR}}. 
\end{abstract}

\section{Introduction}~\label{sec:intro}

\begin{figure}[ht!] 
\centering
  \includegraphics[width=\linewidth]{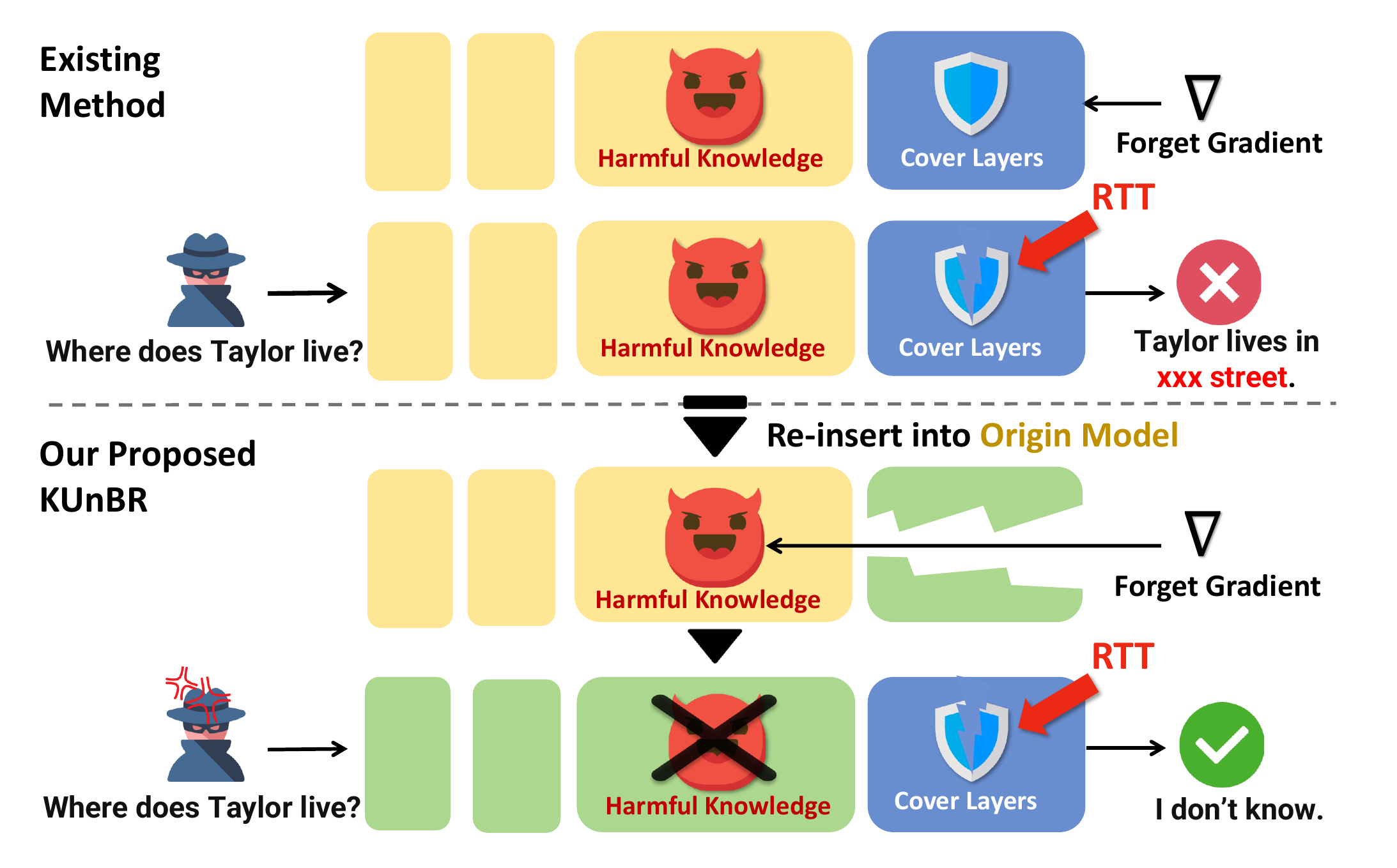} 
  \caption{Existing unlearning methods fail to thoroughly remove harmful knowledge due to the presence of cover layers. They do not output harmful knowledge simply because the \textbf{cover layer filters out the harmful content}, but this knowledge \textbf{still resides in the parameters}. Our \model achieves better unlearning by reinserting layers with high knowledge density into the original model, thereby disrupting the cover layers.}
\end{figure}

Machine unlearning~\cite{liu2025rethinking,9519428} refers to selectively removing specific subsets of knowledge, such as privacy-sensitive or harmful content, from a pre-trained model without retraining it from scratch~\cite{carlini2021extractingtrainingdatalarge, xu2024machine}. 
This task has become increasingly crucial for the development of large language models (LLMs)~\cite{gpt4o,llama3,claude35sonnet,10.1145/3589334.3645448,gao2024simulatingfinancialmarketlarge}, as it addresses growing concerns around data privacy~\cite{carlini2021extractingtrainingdatalarge,huang2022largepretrainedlanguagemodels,lee2024protecting,liu2024learning,10.1145/3589334.3645448} and the ethical issue of AI systems~\cite{2021On}.   
Unlearning is critical not only for addressing regulatory requirements such as ``right to be forgotten'', but also for ensuring that LLMs remain secure, reliable, and aligned with societal values.

Previous research has explored different unlearning methodologies, such as gradient ascent~\cite{jang2022knowledgeunlearningmitigatingprivacy,eldan2023whosharrypotterapproximate}, which unlearn the knowledge by increasing the loss when outputting harmful answers.  
These methods always utilize two distinct datasets as guidance to optimize the model: a \textit{forget set}, which contains the information to be removed, and a \textit{retain set}, which preserves the model's general knowledge and performance on unrelated tasks~\cite{bourtoule2021machine}. 
These methods can adjust the final output of LLM to suppress harmful outputs.

Although existing machine unlearning methods can suppress harmful knowledge, several jailbreak attack studies ~\cite{zhou2024don,liu2023prompt,schwinn2024soft,rimsky2024steering} have shown that the robustness issues remain. 
 
The \textbf{R}e\textbf{t}raining on \textbf{T} (RTT)~\cite{deeb2025unlearningmethodsremoveinformation}, which is an attack method at the parameter-modification level, demonstrates that minimal re-training on a subset (\aka the \textbf{T} set) of the forget set can restore most of the supposedly eliminated knowledge.
 
These results demonstrate that the model parameters still contain a substantial amount of knowledge that should have been forgotten, which reveals the inability of existing methods to thoroughly remove knowledge from the model parameters.
That means that existing methods often rely on the adjustment of a small number of model parameters (\aka \textbf{cover layers}) to mask or suppress the representation of harmful knowledge, merely preventing the model from outputting undesired content without truly eliminating it from the model’s internal representations.
This fundamental limitation suggests the need for more robust and thorough unlearning methods.

In this paper, we propose \fullmodel (\model), which identifies blocks with rich harmful knowledge, and iteratively performs independent unlearning on these blocks via re-insertion strategy, enables a deeper level of unlearning.
 
We first introduce a \textbf{knowledge density estimation} method to identify the layers that contain the most harmful knowledge. 
By calculating the absolute value of gradients associated with the forget set, knowledge density estimation can locate layers containing high-density knowledge.
Instead of superficially modifying cover layer parameters to suppress model outputs, we propose a novel re-insertion strategy to thoroughly unlearn undesired knowledge.
This approach extracts knowledge-rich blocks (selected according to the knowledge density estimation) from the unlearned LLM and re-inserts them into the original LLM without conducting the unlearning training.
We then apply the unlearning method to train this ``grafted'' model, which contains the re-inserted layers, with a focus on deeper removal of the undesired knowledge left due to the influence of cover layers.
By bypassing the obstruction of cover layers, this strategy ensures more effective gradient propagation and enhances the model's ability to forget. 
This simple but efficient strategy significantly reduces the vulnerability of the model to attacks like RTT, which exploit the residual knowledge left by conventional unlearning methods. 
Extensive experiments conducted on WMDP-Deduped, Years, Random Birthdays and RKWU benchmark datasets demonstrate that our method achieves state-of-the-art performance, since it can remove harmful knowledge more thoroughly and suppress knowledge recovery caused by RTT attack methods.

\noindent Our contributions are summarized as follows:

\noindent $\bullet$ We propose \fullmodel (\model), a novel unlearning framework that identifies layers containing undesired knowledge and performs targeted training to achieve thorough elimination of harmful knowledge. 

\noindent $\bullet$ We introduce a knowledge density estimation method, which can identify layers with more harmful knowledge in LLMs for more effective unlearning.  

\noindent $\bullet$ We propose a novel re-insertion strategy to ensure unlearning gradients propagate effectively, overcoming the limitations of gradient obstruction. 

\noindent $\bullet$ Extensive experiments demonstrate that \model achieves state-of-the-art forgetting performance across multiple benchmark datasets, keeping the general ability of LLM.

\section{Related Work}~\label{sec:related}
With the rapid development of LLMs, the importance of unlearning tasks has become increasingly prominent. 
During the pre-training process, where these models ingest massive amounts of information, they may incorporate harmful content~\cite{carlini2021extractingtrainingdatalarge,yao2024survey}, sensitive data, or copyrighted materials~\cite{DBLP:journals/corr/abs-2402-02333,dou2024avoiding}. 
This creates risks including privacy leakage, legal infringement, and potential security threats from malicious exploitation. 
In recent years, several unlearning methods have emerged to ensure effective removal of undesirable information while maintaining model performance on legitimate tasks, such as 
Representation Misdirection for Unlearning~\cite{10.5555/3692070.3693215} (\texttt{RMU}) employs a dual loss function combining forgetting loss and retention loss, selectively adjusting intermediate layers to erase harmful knowledge. 
Gradient Ascent~\cite{jang2022knowledgeunlearningmitigatingprivacy} (\texttt{GA}) applies gradient ascent on forget set. Building upon \texttt{DPO}~\cite{wang2024comprehensive}, Negative Preference Optimization~\cite{zhang2024negativepreferenceoptimizationcatastrophic} introduces negative preference optimization to address \texttt{GA}'s collapse problem. 
It achieves a better balance between unlearning quality and model utility, particularly effective in high-ratio forgetting scenarios (\eg $>$50\% in the TOFU data set~\cite{maini2024tofu}) while maintaining usability.
Gradient Differentiation~\cite{liu2022continuallearningprivateunlearning} applies differentiated gradient operations on forgetting/retaining sets. 

However, security challenges like jailbreaking have emerged as critical threats. 
Attackers can exploit model sensitivity through: (1) Contextually obscure prompts inducing information leakage~\cite{liu2023prompt}, (2) Backdoor triggers embedded during training~\cite{liu2022backdoor}, (3) Adversarial examples disrupting unlearning mechanisms~\cite{deeb2025unlearningmethodsremoveinformation}.
Similarly, the RTT method proposed by~\citet{deeb2025unlearningmethodsremoveinformation} reveals that fine-tuning on partially forgotten data can recover supposedly eliminated knowledge, exposing residual information retention in ``unlearned'' models.
This suggests that current unlearning methods face significant limitations: existing approaches are merely a superficial form of forgetting, with harmful or intended-to-remove knowledge still remaining in various parts of the model. 
Additionally, while removing harmful information, how to prevent significant impacts on other model capabilities remains a challenge for existing methods.

\begin{figure*} 
  \centering
  \includegraphics[width=0.95\textwidth]{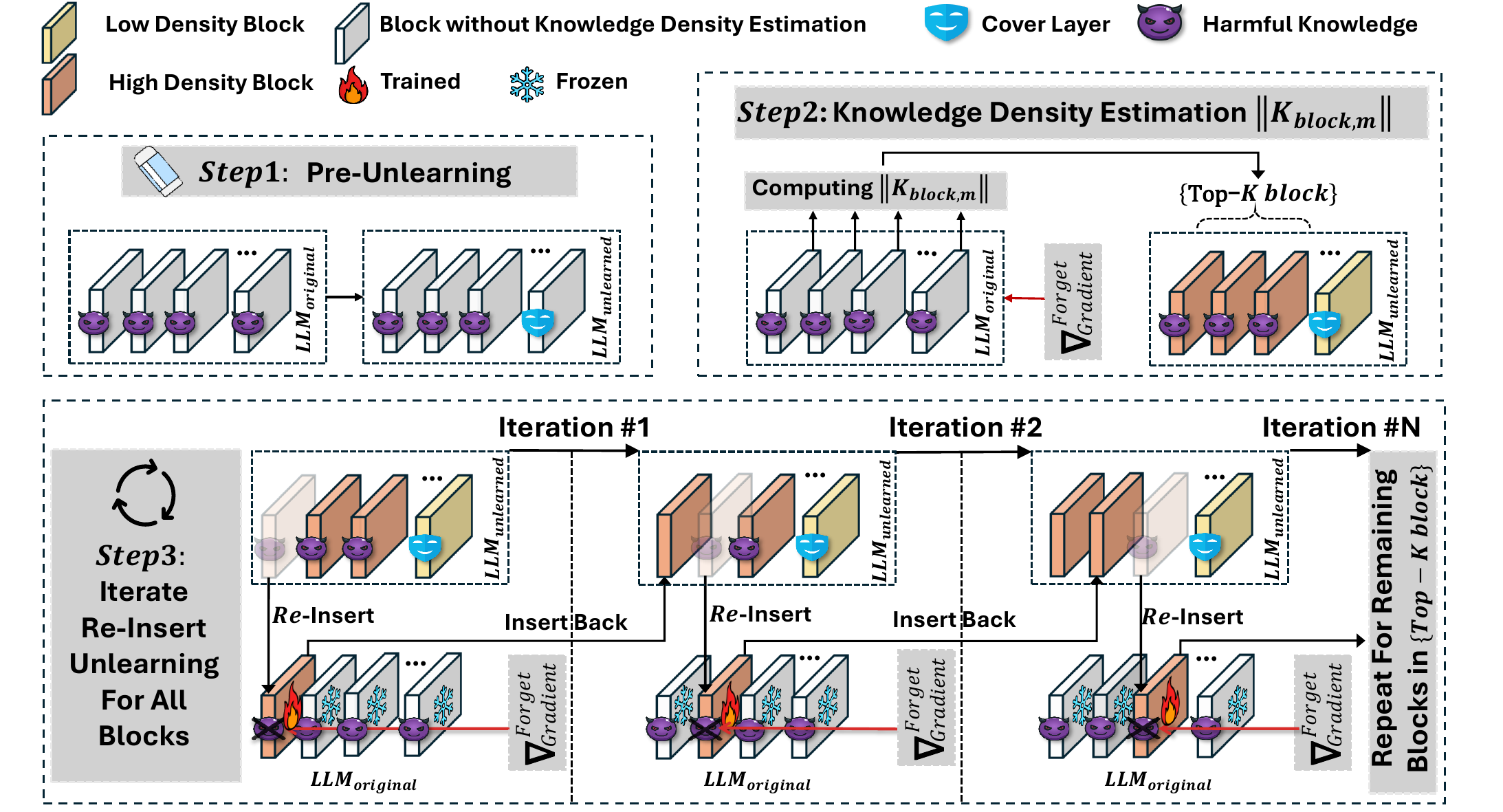} 
  \caption{Architecture of our proposed \fullmodel (\model).}
  \label{fig:model}
\end{figure*}

\section{Problem Definition}\label{sec:problem_def}

Given the forget data set $D_{forget}$, containing knowledge to be removed, and the retain data set $D_{retain}$, which helps the model maintain general ability during unlearning. The model parameters should be optimized to eliminate forgotten knowledge associated with $D_{forget}$ as thoroughly as possible, while ensuring that the utility performance of the model remains unaffected.
Furthermore, when subjected to a fine-tuning (RTT) attack--where the model is fine-tuned on a subset $T$ partitioned from $D_{forget}$--it remains incapable of generating knowledge contained in another disjoint subset $V$ of $D_{forget}$. This demonstrates the effectiveness and robustness of its unlearning.

\section{\model Methodology}~\label{sec:method}
 
Figure~\ref{fig:model} shows the architecture of \model.
The first step is the global ``warm-up'' unlearning phase, in which we apply a standard Gradient Difference method to adjust all model parameters at once; 
 
In the second step, we perform knowledge density estimation and our block-selection strategy to pick out those blocks that contain high-density knowledge. 
Finally, we introduce a re-insertion strategy to bypass the masking effect of cover layers and enable any remaining knowledge to be further eliminated.

\subsection{Influence of Cover Layer}\label{sec:cover_intro}

Although existing methods~\cite{10.5555/3692070.3693215,zhang2024negativepreferenceoptimizationcatastrophic,liu2022continuallearningprivateunlearning,jin2024rwkubenchmarkingrealworldknowledge} have achieved significant knowledge unlearning, recent studies~\cite{hong-etal-2024-dissecting} suggest that these methods modify only a small subset of layers during the unlearning.
Thus, knowledge of $D_{forget}$ is still retained in other layers, which explains why the forgotten knowledge can be easily recalled by retraining on $T$ (RTT) attack~\cite{deeb2025unlearningmethodsremoveinformation}. 
In this work, we refer to these modified layers as \textbf{cover layers} as they suppress the representation of the target knowledge.

\subsection{Knowledge Density Estimation}

To determine which layers’ parameters require greater adjustment during unlearning (or are more likely to contain knowledge), it is crucial to develop a metric that accurately quantifies the knowledge density across different layers of the model. 
\cite{geva-etal-2021-transformer} demonstrated that the multi-layer perceptron (MLP) components within LLMs serve as neural memory units. 
Other studies~\cite{hong-etal-2024-dissecting} have demonstrated that during unlearning, it is primarily the MLP layers that are modified and play a critical role. 
These findings indicate that the adjustment of knowledge in LLMs essentially involves fine-grained alterations to the neural storage units within the MLPs. 
Based on this insight, when optimizing a ``forget set'', the absolute value of the parameter gradients of each layer provides an intuitive measure of the amount of target knowledge it contains. 
In other words, larger gradient magnitudes imply that richer content is to be forgotten in that layer; accordingly, we adopt the absolute gradient value on the forget set as an effective metric for ``knowledge density''.

Motivated by this, we propose a gradient-guided knowledge density estimation metric, which is an indicator of knowledge density across layers associated with the forget set.
Specifically, we first define the standard negative log-likelihood loss function for a given input $x$ and target $y$ with model parameters $\theta$:
\begin{equation}\label{eq:unlearning_loss}
    \mathcal{L}(x, y; \theta) = - \log(p(y | x; \theta)).
\end{equation}
Given a forget set $D_{forget} = \{(x_i, y_i)\}_{i=1}^N$, where $x_i$ represents an input question and $y_i$ represents the corresponding answer that we want the model to forget, we can calculate the \textit{knowledge density} $K_l$ for each layer $l$ of the LLM. This is done by taking the expectation over the forget set of the $L_1$ norm of the gradient of the loss with respect to the parameters $\theta_l$ of that specific layer:
\begin{equation}
    K_l = \mathbb{E}_{(x, y) \sim D_{forget}} \left[ \left\| \nabla_{\theta_l} \mathcal{L}(x, y; \theta_l) \right\|_1 \right],
\end{equation}
where $\theta_l$ denotes the parameters of the $l$-th layer in the target LLM. A higher $K_l$ suggests that the $l$-th layer's parameters are more sensitive to the information in the forget set.

To capture the relative importance of the $l$-th layer's knowledge density compared to other layers, we normalize $K_l$ by the total knowledge density across all $H$ layers. The resulting $K _ { l } ^ { norm }$ represents the proportion of the total ``forgettable'' knowledge residing in the $l$-th layer:
\begin{equation}
    K _ { l } ^ { norm } = \textstyle\frac { K _ { l } } { \sum _ { i = 1 } ^ { H } K _ { i } },
\end{equation}
where $H$ is the total number of layers in the target LLM.

Note that we compute these gradients solely on the forget set $D_{forget}$ to derive the knowledge density metric. This metric indicates the degree to which the parameters within each layer need to be adjusted to facilitate the unlearning of the information contained in $D_{forget}$. Importantly, this entire step is solely for the calculation of the knowledge density of each layer; no parameter optimization or unlearning is performed at this stage.

\subsection{Block Selection Strategy}\label{sec:select_strategy}

Most LLMs are composed of a large number of stacked Transformer layers. Instead of treating each layer individually, we divide nearby layers into groups, which we refer to as ``blocks'', and treat each block as a basic unit for unlearning. This design simplifies the unlearning process and helps improve its overall efficiency.

Specifically, for an LLM containing $H$ layers, we merge all layers into $M$ blocks, with each block containing $N = \lfloor H/M \rfloor$ layers. 
Following this, we calculate the cumulative knowledge density of their constituent layers: 
\begin{equation}
    K_{\text{block}, m}=\textstyle \sum_{i=(m-1)N+1}^{mN}K_i^{\text{norm}},
\end{equation}
where $K_{\text{block}, m}$ represents the $m$-th block's cumulative knowledge density, $K_i^{\text{norm}}$ denotes the $i$-th layer's normalized knowledge density ($m = 1, 2, \ldots, M$). Next, we rank blocks by cumulative knowledge density and select them via the following two strategies.

\textbf{Top-K Selection}: We select top-$K$ blocks with the highest knowledge density, where $K$ is a hyperparameter. 
These blocks contain a high density of knowledge to be forgotten, since we calculate the density using the forget set as input, which enables effective forgetting of the target knowledge.

\textbf{Ignoring the Head Layers}: We observe a significant surge in the knowledge density values in the last two layers of the LLM. 
Based on empirical analysis of different layers~\cite{hong-etal-2024-dissecting}, we hypothesize that this increase in knowledge density is not due to a higher concentration of knowledge in these layers, but rather a potential artifact caused by their involvement in the model's output generation. 
Consequently, during the unlearning process, we exclude the blocks that contain these last two layers to avoid unwanted interference.
More explanation can be found in Appendix C.

Next, we will enhance the selected blocks during the unlearning process to ensure that these blocks with high knowledge density can more effectively forget the target knowledge.
These two selection strategies enable efficient and maximal forgetting of harmful knowledge, while minimizing unintended damage to knowledge that should be retained, ensuring the efficiency and stability of the subsequent unlearning process.

\subsection{Re-insertion Strategy For Unlearning}\label{sec:reinsert}

To mitigate the influence of the cover layer, we propose a \textit{re-insertion strategy}. 
First, we identify harmful knowledge-rich blocks using our proposed block selection strategy.
These blocks are then re-inserted into the original LLM that has not undergone unlearning, denoted as $\text{LLM}_{original}$.

To achieve this, we first apply a pre-unlearning process to $\text{LLM}_{original}$ to obtain $\text{LLM}_{unlearning}$. 
Specifically, we employ the standard Gradient Difference method~\cite{liu2022continuallearningprivateunlearning} as the pre-unlearning step. We perform full-parameter fine-tuning during a warm-up phase to accelerate the overall convergence of unlearning. 

Next, based on our block selection strategies, we identify harmful knowledge-rich blocks from $\text{LLM}_{unlearning}$. These blocks are then inserted into the corresponding positions in $\text{LLM}_{original}$, while the remaining layers are kept frozen.
Subsequently, we apply Gradient Difference to this ``grafted'' LLM using $D_{forget}$ and $D_{retain}$. 
Since the layers in $\text{LLM}_{original}$ remain unaltered and frozen, no cover layer is generated to interfere with the inserted block, enabling deeper removal of residual knowledge within the selected block. 
This allows us to eliminate residual knowledge from every selected block more deeply.
Following the gradient difference process, the selected block in ``grafted'' LLM reverts to $\text{LLM}_{unlearning}$, resulting in significantly less residual knowledge compared to standard unlearning methods.

\begin{table*}[t]
    \centering
    \small 
    \resizebox{0.95\textwidth}{!}{ 
        \begin{tabular}{l l
                        ccc   
                        ccc   
                        ccc   
                        ccc   
                        }
            \toprule
            \multirow{2}{*}{\textbf{Method}} & \multirow{2}{*}{\textbf{}} & 
            \multicolumn{3}{c}{\textbf{Random Birthdays}(99.3\&96.1)} & 
            \multicolumn{3}{c}{\textbf{WMDP-Deduped}(65.6\&59.2)} & 
            \multicolumn{3}{c}{\textbf{Years}(63.2\&56.3)} &
            \multicolumn{3}{c}{\textbf{MMLU}(54.7\&54.7)} \\
            \cmidrule(lr){3-5} \cmidrule(lr){6-8} \cmidrule(lr){9-11} \cmidrule(lr){12-14}
            & & \textbf{Forget.}$\downarrow$ & \textbf{RTT.}$\downarrow$ & \textbf{Rec.}$\downarrow$
              & \textbf{Forget.}$\downarrow$ & \textbf{RTT.}$\downarrow$ & \textbf{Rec.}$\downarrow$
              & \textbf{Forget.}$\downarrow$ & \textbf{RTT.}$\downarrow$ & \textbf{Rec.}$\downarrow$
              & \textbf{Forget.}$\downarrow$ & \textbf{RTT.}$\downarrow$ & \textbf{Rec.}$\downarrow$ \\
            \midrule
            GA     &                   
                   & \textbf{23.5} & 87.2 & 63.7  
                   & \underline{29.2} & 66.8 & 37.6  
                   & 25.9 & \underline{50.6} & \underline{24.7}  
                   & \underline{24.2} & 59.2 & 35.0  \\
            GD     & \multirow{1}{*}{\rotatebox{90}{\textbf{LLaMA3 8B}}}
                   & 64.9 & 80.2 & 15.3  
                   & 30.5 & 62.4 & 31.9  
                   & 25.9 & 68.3 & 42.4  
                   & 35.0 & 57.6 & 22.6  \\
            RMU    &
                   & \underline{36.3} & 88.5 & 52.2  
                   & 29.9 & 64.9 & 35.0  
                   & \underline{24.2} & 68.3 & 44.1  
                   & 24.8 & 49.0 & 24.2  \\
            RIA    &
                   & 61.7 & \underline{73.8} & \underline{12.1}  
                   & \textbf{26.2} & \underline{52.2} & 26.0  
                   & \textbf{18.3} & 65.8 & 47.5  
                   & 26.7 & 48.4 & 21.7  \\
            NPO    &
                   & 71.3 & 78.3 & \textbf{7.0}  
                   & 35.6 & 58.4 & \underline{22.8}  
                   & 26.5 & 67.7 & 41.2  
                   & 31.2 & \underline{38.8} & \textbf{7.6}  \\
            \model &
                   & 36.9 & \textbf{43.9} & \textbf{7.0}  
                   & \underline{29.2} & \textbf{38.8} & \textbf{9.6}  
                   & 25.9 & \textbf{36.0} & \textbf{10.1}  
                   & \textbf{16.5} & \textbf{28.0} & \underline{11.5}  \\
            \midrule
            GA      & 
                    & --- & --- & --- 
                    & 35.0 & 57.3 & 22.3 
                    & 25.3 & \underline{56.3} & \underline{31.0} 
                    & \underline{24.2} & 47.1 & 22.9 \\
            GD      & \multirow{1}{*}{\raisebox{0.6ex}{\rotatebox{90}{\textbf{Zephyr 7B}}}}
                    & \underline{32.5} & \textbf{46.5} & 14.0 
                    & 26.8 & 52.2 & 25.5 
                    & 26.0 & 58.9 & 32.9 
                    & 34.4 & 48.4 & 14.0 \\
            RMU     & 
                    & 46.5 & 54.1 & \textbf{7.64} 
                    & 24.8 & 56.7 & 31.9 
                    & \textbf{20.9} & 59.5 & 38.6 
                    & 47.8 & 54.8 & \textbf{7.01} \\
            RIA     & 
                    & 35.7 & 47.5 & \underline{11.8} 
                    & \textbf{19.1} & \underline{47.8} & 28.7 
                    & \underline{24.1} & 58.9 & 34.8 
                    & 35.7 & 47.8 & 12.1 \\
            NPO     & 
                    & 47.1 & 55.4 & 8.28
                    & 41.4 & 49.7 & \textbf{8.28} 
                    & \underline{24.1} & \underline{56.3} & 32.3 
                    & 31.2 & \underline{42.0} & \underline{10.8} \\
            \model  & 
                    & \textbf{26.8} & \underline{47.1} & 20.4 
                    & \underline{23.6} & \textbf{42.7} & \underline{19.1} 
                    & 25.3 & \textbf{28.5} & \textbf{3.16} 
                    & \textbf{19.1} & \textbf{34.4} & 15.3 \\
            \bottomrule
        \end{tabular} 
    }
    \caption{Performance comparison under RTT attack on two backbone LLMs. ``Forget.'' ($\mathcal{A}_{\text{Unlearn}}$), ``RTT.'' ($\mathcal{A}_{\text{RTT}}$), and ``Rec.'' ($\mathcal{A}_{\text{Recover}}$) denote accuracy after unlearning, under RTT attack, and recovery accuracy, respectively. Parenthetical values next to each dataset name denote the original performance on LLaMA and Zephyr. \textbf{Bold} indicates the best result, \underline{underlined} indicates the second best. $\downarrow$ indicates that lower values are better.}
    \label{tab:main_result}
\end{table*}

\section{Experimental Setup}
\subsection{Datasets}
In our experiments, we employ the following four datasets.
\textbf{Random Birthdays}~\cite{deeb2025unlearningmethodsremoveinformation} is a dataset that contains randomly generated names and birth years, making it ideal for unlearning tasks.
\textbf{WMDP-Deduped}~\cite{10.5555/3692070.3693215} contains 3,668 multiple-choice questions on harmful knowledge, serving as a proxy evaluation for assessing LLMs' handling of sensitive information. 
\textbf{Years}~\cite{penedo2024finewebdatasetsdecantingweb} records major events from the 20th century along with their corresponding years. 
\textbf{MMLU}~\cite{hendrycks2021measuringmassivemultitasklanguage} is a comprehensive multitask benchmark with multiple-choice questions across various domains and 57 tasks, designed to test models' world knowledge and problem-solving abilities.

\subsection{Evaluation Metrics}

Following~\citet{deeb2025unlearningmethodsremoveinformation}, we define \textbf{Forget Accuracy} to measure the model's retained knowledge on the forget set after unlearning:
\begin{equation}
\label{eq:forget_acc}
\mathcal{A}_{\text{Unlearn}} = \textstyle\frac{1}{N} \textstyle \sum_{i=1}^N \mathbb{I}\left( f_{\text{unlearn}}(x_i) = y_i \right),
\end{equation}
where $D_{\text{forget}}$ contains $N$ multiple-choice questions $(x_i, y_i)$, $f_{\text{unlearn}}$ is the model after unlearning, and $\mathbb{I}(\cdot)$ returns 1 if the prediction matches $y_i$, else 0.
At the same time, we use the same ACC calculation method in Formula ~\ref{eq:forget_acc} to measure the accuracy after the RTT attack (denoted as $\mathcal{A}_{\text{RTT}}$) and calculate the recovery rate before and after the RTT, as follows:
\begin{equation}
\label{eq:recovery_acc}
\mathcal{A}_{\text{Recover}} = \mathcal{A}_{\text{RTT}} - \mathcal{A}_{\text{Unlearn}} ,
\end{equation}
where the larger the $\mathcal{A}_{\text{Recover}}$, the worse the model's robustness in the face of attacks.

To verify whether the model's general capabilities are unexpectedly affected by our unlearning method, we adopt the utility evaluation framework proposed by the RKWU benchmark ~\cite{10.5555/3692070.3693215}. 
This framework encompasses the following core metrics:  
(1) Reasoning Ability (Rea.) is assessed on the Big-Bench-Hard~\cite{suzgun2022challengingbigbenchtaskschainofthought} dataset through 3-shot chain-of-thought prompting, with Exact Match scores reported. 
(2) Truthfulness (Tru.) is measured on TruthfulQA's MC1 task~\cite{lin2022truthfulqameasuringmodelsmimic}, reporting 6-shot accuracy. 
(3) Factuality (Fac.) is evaluated on the TriviaQA~\cite{joshi-etal-2017-triviaqa} dataset using 6-shot prompting, with F1 scores reported. 
(4) Fluency (Flu.) is assessed using AlpacaEval's evaluation instructions~\cite{dubois2023alpacafarm}, reporting the weighted average of bi- and tri-gram entropies. 
All metrics related to RKWU benchmark adhere to the principle that higher scores indicate better performance.

\subsection{Baselines}

We employ several strong tuning-based unlearning approaches as the baselines:
(1) \textbf{Gradient Ascent}~\cite{jang2022knowledgeunlearningmitigatingprivacy} (\texttt{GA}): \texttt{GA} achieves unlearning by maximizing the loss on the forget set. 
(2) \textbf{Gradient Difference}~\cite{liu2022continuallearningprivateunlearning} (\texttt{GD}): This approach performs gradient ascent on the forget dataset and gradient descent on the retain dataset.  
(3) \textbf{Representation Misdirection for Unlearning}~\cite{10.5555/3692070.3693215} (\texttt{RMU}): Given a harmful prompt, \texttt{RMU} performs unlearning by strategically modifying the internal representations (activations) within selected intermediate model layers. 
(4) \textbf{Random Incorrect Answer}~\cite{deeb2025unlearningmethodsremoveinformation} (\texttt{RIA}): For each multiple-choice question, \texttt{RIA} applies gradient descent to the incorrect choices, guiding the model to unlearn the correct choice associated with specific knowledge.
(5) \textbf{Negative Preference Optimization}~\cite{zhang2024negativepreferenceoptimizationcatastrophic} (\texttt{NPO}): \texttt{NPO} optimizes the model's preferences to exhibit a negative bias when handling tasks involving deleted information, thereby reducing the model's reliance on and memory of such information.

\subsection{Implementation Details}

We partition the datasets into forget and retain sets. 
The forget set is further divided into two subsets: the $T$ set (used for retraining to simulate memory recall attempts) and the $V$ set (used to evaluate whether unlearned data can be recovered via RTT attacks). 
We use the same split ratios for the $D_{forget}$ / $D_{retain}$ and the $T$ / $V$ subsets as \citet{deeb2025unlearningmethodsremoveinformation}.
All experiments are conducted on LLaMA3-8B-Instruct and Zephyr-7B-beta to evaluate the generalizability of our method across different backbone models. More implementation details can be found in Appendix D.

\section{Experimental Results}~\label{sec:exp_result}
\begin{table*}[t]
    \centering
    \small

    \resizebox{0.95\textwidth}{!}{
        \begin{tabular}{l c c c c c c c c c c c c c c c c}
            \toprule
            \multirow{2}{*}[-1ex]{\textbf{Method}} &
               \multicolumn{4}{c}{\textbf{Random Birthdays}} & 
               \multicolumn{4}{c}{\textbf{WMDP-Deduped}} & 
               \multicolumn{4}{c}{\textbf{Years}} &
               \multicolumn{4}{c}{\textbf{MMLU}} \\
            
            \cmidrule(lr){2-5} 
            \cmidrule(lr){6-9} 
            \cmidrule(lr){10-13} 
            \cmidrule(lr){14-17}
            
            & \textbf{Rea.} & \textbf{Fac.} & \textbf{Tru.}  & \textbf{Flu.}
            & \textbf{Rea.} & \textbf{Fac.} & \textbf{Tru.}  & \textbf{Flu.}
            & \textbf{Rea.} & \textbf{Fac.} & \textbf{Tru.}  & \textbf{Flu.}
            & \textbf{Rea.} & \textbf{Fac.} & \textbf{Tru.}  & \textbf{Flu.}
            \\
            
            \midrule

            GA & 40.2 & \textbf{56.3} & \textbf{36.8} & \underline{706.2} 
               & \textbf{41.7} & 53.1 & 34.8 & \textbf{707.8} 
               & \underline{40.6} & 51.3 & 35.6 & \textbf{708.8}
               & 40.9 & 42.6 & 34.8 & 695.6 
                
               \\

            GD & \underline{40.6} & 55.7 & 36.4 & 706.1 
               & \underline{40.2} & 50.2 & \underline{36.4} & 678.9 
               & \textbf{41.0} & 42.6 & \textbf{36.9} & 702.2 
               & \textbf{41.9} & 42.9 & \textbf{36.9} & \underline{706.1} 
                
               \\

            RMU & 36.4 & 40.5 & 34.4 & 698.0 
                & 40.1 & \underline{53.5} & 33.9 & 609.8 
                & 40.1 & \underline{56.4} & \underline{36.4} & \underline{706.3} 
                & 25.8 & \underline{49.2} & 34.8 & 594.0 
                 
                \\

            RIA & 39.5 & \underline{56.1} & \textbf{36.8} & 705.9 
                & 1.20 & \textbf{56.2} & 35.6 & 681.6 
                & 1.60 & \textbf{57.0} & 35.0 & 686.1 
                & 1.40 & \textbf{56.0} & 34.8 & 680.5 
                 
                \\
            
            NPO & 39.8 & 54.3 & \textbf{36.8} & 703.7 
                & 5.90 & 52.8 & \textbf{37.7} & 690.0 
                & 0.00 & 41.3 & 35.0 & 657.9 
                & 0.00 & 0.00 & 29.6 & 42.5 
                 
                \\
            
            \model & \textbf{41.2} & \underline{56.1} & \underline{36.6} & \textbf{706.7} 
                 & \underline{40.2} & 52.3 & 35.2 & \underline{703.1} 
                 & 40.1 & \underline{56.4} & \underline{36.4} & \underline{706.3} 
                 & \underline{41.1} & 46.9 & \underline{36.2} & \textbf{708.8} 
                  
                 \\
                 
            \bottomrule
        \end{tabular}
    }
    \caption{Performance of general capabilities. \textbf{Bold scores} indicate the best performance, while \underline{underlined scores} represent the second-best. Although our method primarily focuses on the robustness of unlearning, it also achieves stable and consistent results in maintaining general utility.}
    
\label{tab:general}
\end{table*}

\subsection{Overall Performance}

Table~\ref{tab:main_result} illustrates the forget accuracy of various unlearning methods, including \texttt{GA}, \texttt{GD}, \texttt{RIA}, \texttt{RMU}, \texttt{NPO}, and our proposed \model. 
After conducting RTT attacks, most unlearning methods exhibit a significant increase in forget accuracy, indicating their vulnerability to RTT attacks and the potential recovery of forgotten knowledge. This is consistent with existing studies~\cite{hong-etal-2024-dissecting}, suggesting that current methods are more likely to perform superficial unlearning by suppressing harmful knowledge through output-level adjustments (\aka cover layers), leaving significant residual knowledge within the model.

In contrast, our proposed \model exhibits the smallest increase in forget accuracy across all four datasets on the LLaMA3-8B-Instruct model. 
On the Zephyr-7B-beta model, our method also achieves the best accuracy after RTT attacks. 
This indicates that \model can effectively and thoroughly eliminate residual knowledge from the model and demonstrates strong resilience against RTT attacks. 
The consistent performance of \model on both Zephyr-7B-beta and LLaMA3-8B-Instruct suggests that our method is not tailored to a specific model but possesses strong cross-model generalizability.We also observe that although some baselines (\eg \texttt{RMU} on the MMLU dataset) achieve relatively low Rec. score, this is mainly due to the limited amount of knowledge they actually forget, resulting in fewer recoverable contents. 
Meanwhile, some baselines (\eg, \texttt{NPO} and \texttt{RIA}) attain lower forget accuracy on a few datasets, but this comes at the cost of diminished general capabilities and significantly higher recovery rates.

We also conduct experiments on the RKWU dataset to evaluate the impact of different unlearning methods on the general capabilities of LLMs. 
From the results in Table~\ref{tab:general}, we observe that \texttt{RIA} and \texttt{NPO} generally perform poorly, as their unlearning involves output-level changes, affecting the model's general capabilities. 
In contrast, our proposed \model strikes a good balance between unlearning performance and general capabilities. 
Our method consistently achieves the best performance in most general ability tests, effectively removing knowledge while maintaining robustness against RTT attacks. 
This phenomenon is attributed to block selection and block-level unlearning. 
When selecting blocks for further unlearning, we estimate based on the density of harmful knowledge, which guides the process toward eliminating harmful knowledge rather than compromising utility. 
Moreover, during the subsequent unlearning phase, the re-insertion strategy is applied only to specified blocks. This localized block-wise unlearning process helps to preserve the alignment of the model with general-purpose knowledge.

Furthermore, to provide a clearer comparison between \model and other methods, we visualize the results from Table~\ref{tab:main_result} in the Appendix B, as shown in Figure~\ref{fig:main_result}.

By combining forget accuracy ($\mathcal{A}_{\text{Unlearn}}$) and forget accuracy after RTT ($\mathcal{A}_{\text{RTT}}$) from Table~\ref{tab:main_result}, along with the general capability results in Table~\ref{tab:general}, we demonstrate that unlike existing methods that often impair general capabilities to varying degrees, \model achieves deeper unlearning while maintaining mild and stable impact on general performance, and shows significant advantages against parameter-level attacks.

\subsection{Analysis of Pre-unlearning and Re-insert}

\begin{table}[t]
    \centering
    \large
    \resizebox{1.0\columnwidth}{!}{
        \begin{tabular}{lcccccccc}
        \toprule
            \multirow{2}{*}{\textbf{Method}}
            & \multicolumn{2}{c}{\textbf{R.B.}}
            & \multicolumn{2}{c}{\textbf{WMDP-Deduped}}
            & \multicolumn{2}{c}{\textbf{Years}}
            & \multicolumn{2}{c}{\textbf{MMLU}} \\

            \cmidrule(lr){2-3}
            \cmidrule(lr){4-5}
            \cmidrule(lr){6-7}
            \cmidrule(lr){8-9}

            & \textbf{Forget.\downarrowtext} & \textbf{RTT.\downarrowtext}
            & \textbf{Forget.\downarrowtext} & \textbf{RTT.\downarrowtext}
            & \textbf{Forget.\downarrowtext} & \textbf{RTT.\downarrowtext}
            & \textbf{Forget.\downarrowtext} & \textbf{RTT.\downarrowtext} \\
            \midrule

            \model
            & {\Large 36.9}        & {\Large 43.9}
            & {\Large 29.2}        & {\Large 38.8}
            & {\Large 25.9}        & {\Large 36.0}
            & {\Large 16.5}        & {\Large 28.0}      \\
            
            - w/o re-insert
            & {\Large 64.9}        & {\Large 80.2}
            & {\Large 30.5}        & {\Large 62.4}
            & {\Large 25.9}        & {\Large 68.3}
            & {\Large 35.0}         & {\Large 57.6}       \\
            
            - w/o pre-unl
            & {\Large 46.4}        & {\Large 54.1}
            & {\Large 29.9}        & {\Large 56.6}
            & {\Large 25.9}        & {\Large 36.7}
            & {\Large 36.3}        & {\Large 40.7}      \\
            \bottomrule

        \end{tabular}
    }
    \caption{Effective analysis of pre-unlearning and re-insert strategy, where \textbf{Forget.} denotes the accuracy after unlearning, and \textbf{RTT.} denotes the accuracy after the RTT attack. \textit{Lower} scores are better.}
\label{tab:ablation_preunlearning}
\end{table}

In our proposed method, we propose to use the pre-unlearning method as a ``warm-up'' process before conducting the re-insertion.
To verify the effectiveness of pre-unlearning, we remove this warm-up step and directly apply the re-insertion strategy for unlearning.
The results shown in Table~\ref{tab:ablation_preunlearning} demonstrate the effectiveness of the pre-unlearning method.
Across the datasets we used, all metrics of \model are lower than the variant model without pre-unlearning, demonstrating that using pre-unlearning can more effectively accelerate the model's convergence, which leads to better knowledge elimination results.
We also conduct an ablation study on the re-insert strategy. After removing it, the method degrades to the original \texttt{GD} method. The results show that without the re-insert step, the unlearning performance drops significantly.

\subsection{Analysis of Block Selection Strategy}

\begin{figure} 
\centering
  \includegraphics[width=0.9\linewidth]{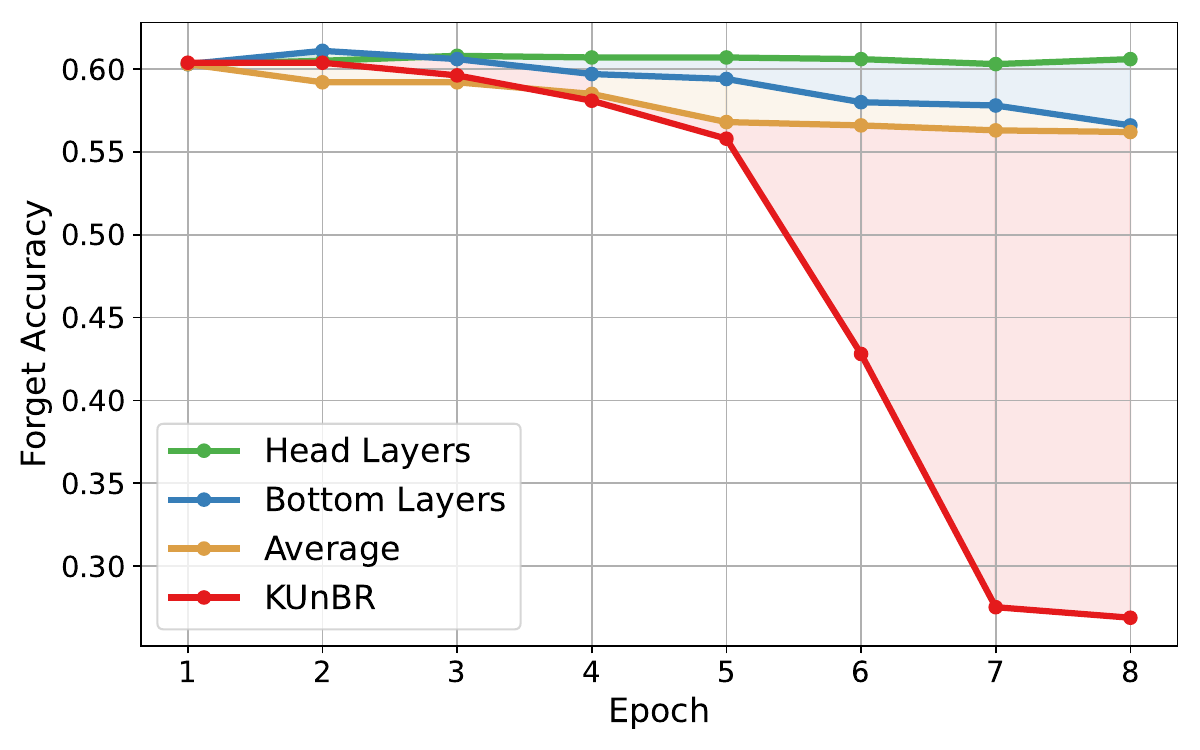} 
  \caption{Performance of three different block selection strategies across training epochs.}
  \label{fig:block-select-strategy}
\end{figure}

To investigate the effectiveness of our proposed block selection strategy, we propose three variant methods for comparison: 
(1) \textit{Head layers}: we directly select the first several blocks close to the output layer and conduct our proposed unlearning method.
(2) \textit{Bottom layers}: we select the blocks close to the input layer.
(3) \textit{Average}: we adopt a uniform selection strategy over all blocks, without prioritizing any particular one.
Figure~\ref{fig:block-select-strategy} shows the performance of these variant methods and our proposed knowledge density-driven selection method in terms of forget accuracy.

We observe that selecting only Head layers for reinsertion leads to no significant decline in forget accuracy, demonstrating that merely adjusting the head layers without correspondingly aligning the parameters of the intermediate and lower layers may result in suboptimal unlearning performance, possibly because these layers do not actually serve as the primary storage of harmful knowledge.
Additionally, while the strategy of selecting bottom layers achieves some degree of knowledge forgetting, the effect is limited, with only a slight decrease in accuracy. Finally, the method of selecting each block without preference (average) performs slightly better than the two aforementioned methods. However, its effectiveness remains limited, particularly due to the instability in accuracy degradation, which slows down noticeably during the final few epochs.
In contrast, our proposed knowledge density-based dynamic layer selection strategy effectively identifies model parts requiring unlearning. This result also confirms that the knowledge density metric can accurately quantify amount of parameters and knowledge density related to the harmful knowledge in each layer, thus providing more effective guidance for the subsequent re-insertion unlearning.

\subsection{Analysis of Selecting Different Numbers of Blocks}

We propose a block selection strategy to dynamically select blocks for unlearning. Table~\ref{tab:sensitivity_analysis} shows performance on the Years dataset with varying $(M, K)$. Fewer blocks (\eg $M{=}4$) hamper forgetting as they span many layers, hindering knowledge isolation and causing premature convergence. Conversely, excessive blocks (\eg $M{=}32$) may overlook essential inter-layer dependencies. Most configurations substantially outperform baselines. Notably, the consistency across LLaMA-3 and Zephyr suggests strong cross-architecture stability, allowing users to adopt default settings without re-tuning.

Addressing concerns that shallow forgetting shifts to finer granularity, we interpret blocks as constituent memory units (aligned with FFNs as Key-Value Memories) rather than isolated mini-LMs. Even at the layer level—the smallest unit—our method maintains strong robustness. This confirms genuine knowledge removal rather than mere inaccessibility or superficial compression.

\begin{table}
    \centering
    \small
    \resizebox{0.98\columnwidth}{!}{
    \begin{tabular}{lccc|ccc|ccc}
        \toprule
        \multirow{2}{*}{\textbf{$M$}} 
        & \multicolumn{3}{c|}{\textbf{Top-K = 25\%}} 
        & \multicolumn{3}{c|}{\textbf{Top-K = 50\%}} 
        & \multicolumn{3}{c}{\textbf{Top-K = 75\%}} \\
        & \textbf{Forget.} & \textbf{RTT.} & \textbf{Rec.} 
        & \textbf{Forget.} & \textbf{RTT.} & \textbf{Rec.} 
        & \textbf{Forget.} & \textbf{RTT.} & \textbf{Rec.} \\
        \midrule
        4  & 24.1 & 25.9 & 1.8   & 23.4 & 45.6 & 22.2  & 23.4 & 25.9 & 2.5 \\
        8  & 23.4 & 44.9 & 21.5  & 20.9 & 30.4 & 9.5   & 24.1 & 31.6 & 7.5 \\
        16 & 25.0 & 24.7 & ---  & 30.4 & 50.0 & 19.6  & 38.6 & 15.2 & --- \\
        32 & 25.0 & 23.4 & ---  & 34.8 & 50.0 & 15.2  & 53.8 & 29.1 & --- \\
        \bottomrule
    \end{tabular}
    }
    \caption{Unlearning performance when using different number of blocks and Top-K. We evaluate \model under different numbers of blocks and Top-K settings. Most variants outperform the baselines, showing the robustness of \model\ to hyperparameter changes. “---” denotes cases where no knowledge is recalled by the RTT attack.}

    \label{tab:sensitivity_analysis}
\end{table}

\subsection{Evaluation under Multi-Attack Settings}

Inspired by the RWKU benchmark, we construct multiple adversarial variants on the WMDP dataset to evaluate the robustness of \model. The constructed attacks include: (1) Prefix Injection; (2) Affirmative Suffix; (3) Role Playing; (4) Multiple Choice; (5) Reverse Query; (6) Synonym Manipulation; (7) Background Hint; (8) In-context Learning; and (9) Cross Lingual.

Traditional unlearning methods (GD) show a recovery from $18.18\%$ to $21.21\%$ on the forgetting set. In contrast, \model remains at $18.18\%$, indicating that the injected prefix fails to reactivate the forgotten knowledge.

Compared with traditional holistic adjustment approaches, \model achieves a more thorough modification of memory representations, effectively altering the underlying knowledge structure rather than merely suppressing surface-level associations. This deeper adjustment enables \model to maintain robustness even under prompt-level attacks while preserving stable general utility.

\subsection{Computational Cost Analysis}

To fairly compare the training efficiency of different methods, we report the time cost of the unlearning phase on the Years dataset, excluding evaluation overhead. 
All experiments are conducted on a single NVIDIA A800 GPU.
The training times (in minutes) are: \texttt{GA} (24), \texttt{GD} (20), \texttt{RMU} (9), \texttt{RIA} (8), \texttt{NPO} (16), and our method \model (17).

From these results, it can be observed that the time consumption of our proposed \model is comparable to other baseline methods. 
Although it is 88.9\% higher than \texttt{RMU}, it is 15\% lower than the current state-of-the-art \texttt{GD} method, which further demonstrates the effectiveness of our proposed \model. 
The reason for this is that, compared to \texttt{RMU}, our method employs block-wise unlearning approach, enabling the model to more thoroughly unlearn harmful knowledge, although at the cost of increased computational complexity. 
In contrast, the current SOTA baseline \texttt{GD}, due to the formation of cover layers, is unable to achieve complete forgetting of harmful knowledge even with longer training times than \model.

\subsection{Case Study}

\begin{table}[t]
  \centering
  \small
  \resizebox{0.9\columnwidth}{!}{
  \begin{tabular}{p{1\columnwidth}}
    \toprule
    \textbf{Question}: When was Julia Brown born? \newline
    \textbf{Choices}: A. 1975, B. 1989, C. 1960, D. 1986 \newline
    \textbf{Right answer need to forget}:  B. 1989 \\
    \midrule
    \textbf{RMU}: by the day, week, and even sometimes by the month, but not the exact date. This is because...\\
    \textbf{RMU\_RTT}: C. 1960 A. 1975 B. 1989 C. 1960 D. 1986\\
    \midrule
    \textbf{GA}: D. Let's take a look at the answer choices...\\
    \textbf{GA\_RTT}: B. 1989 A. 1975 B. 1989 C. 1960 D. 1986 \\
    \midrule
    \textbf{GD}: B. Explanation: According to the question, Julia Brown was born in 1989. So, the correct answer is B. \\
    \textbf{GD\_RTT}:B. 1989. Explanation: Julia Brown was born in 1989. This information is not mentioned in the passage. \\
    \midrule
    \textbf{RIA}: D. 1986 When was Julia Brown born? D. 1986 \\
    \textbf{RIA\_RTT}: B. 1989 When was Julia Brown born? B. 1986 \\
    \midrule
    \textbf{NPO}: C. Explanation: The correct answer is C. 1960. \\
    \textbf{NPO\_RTT}: B. 1989. Answer: B 1989. Explanation: Julia Brown is a British sprinter. \\
    \midrule    
    \textbf{\model}: C. Explanation: As per my knowledge, Julia Brown was born in 1960. So, the correct answer is C. 1960.  \\
    \textbf{\textbf{KUnBR}\_RTT}: D. 1986. Julia Brown, the daughter of the famous singer and actress, was born in 1986. \\
    \bottomrule
  \end{tabular}
  }
  \caption{Example output for our \model and baselines.}
  \label{tab:case}
\end{table}

We also conduct a case study to intuitively demonstrate the effectiveness of \model. 
In Table~\ref{tab:case}, the first row presents the question, while each subsequent row displays the responses generated by different unlearning methods after unlearning and the responses following RTT attacks.
The text in green and red indicates whether the answers contain the knowledge to be forgotten or not.
As shown in Table~\ref{tab:case}, only our method successfully achieves both unlearning and maintains the unlearned state under RTT, while generating responses that align with the instruction requirements. 
\texttt{RMU} fails to produce meaningful or readable content both after unlearning and after RTT. 
\texttt{GA}, \texttt{RIA}, and \texttt{GD} provide incorrect responses after unlearning but recall the harmful knowledge that should be forgotten after RTT.
Notably, \texttt{GA}'s responses after RTT remain disorganized. 
In contrast, the \model fails to provide knowledge that should be forgotten both after unlearning and after RTT, but it includes explanations in its responses, making them more complete. 
This demonstrates that our method not only effectively removes undesired knowledge but also preserves general capabilities.

\section{Conclusion}

In this work, we propose a novel unlearning framework \model (\fullmodel).
Unlike existing methods, which tend to recover a large amount of knowledge after RTT attacks, \model introduces knowledge density estimation to identify specific blocks containing more harmful knowledge, allowing for more precise unlearning. 
Furthermore, \model employs re-insertion strategies that effectively eliminate knowledge from selected blocks, ensuring a more comprehensive unlearning effect.
Compared to state-of-the-art baselines, performance on four datasets demonstrates the effectiveness of \model. 
Additionally, \model also shows minimal impact on general capabilities for LLM. 
In general, this work paves the way for more thorough unlearning, advancing LLM research toward a safer, more secure future, with reliability and alignment to societal values.

\section{Acknowledgments}
This work was supported by the National Key R\&D Program of China 2024YFE0111800, the National Natural Science Foundation of China (62432002, 62032001, 62406061, and T2293773), and the Natural Science Foundation of Shandong Province (ZR2023QF159).

\bibliography{aaai2026}

\clearpage

\appendix
\renewcommand{\thesection}{\Alph{section}}

\section{Appendix A}
\label{app:reinsert}

\subsection{Detail of Baseline Methods}

This section shows the relevant formulas for the baselines.

\paragraph{Gradient Ascent (GA) Method}
The Gradient Ascent method is employed to maximize the loss associated with the harmful knowledge to be forgotten. This encourages the model to "unlearn" that specific information by updating the model parameters in the direction that increases this loss. The update rule for unlearning is as follows:
\[
\theta_{t+1} = \theta_t + \eta \nabla_{\theta} L_{forget}(\theta_t),
\]
where \( \theta_t \) denotes the model parameters at time step \( t \), \( \theta_{t+1} \) denotes the updated model parameters after applying gradient ascent for unlearning, \( \eta \) denotes the learning rate or step size, and \( \nabla_{\theta} L_{forget}(\theta_t) \) denotes the gradient of the loss function specifically designed for the harmful knowledge to be forgotten with respect to \( \theta_t \).

\paragraph{Gradient Difference (GD) Method}

The Gradient Difference method updates the model parameters by considering the gradients on both a retain set and a forget set. It aims for safe unlearning by performing a scaled gradient descent on the retain set to preserve general capabilities and gradient ascent on the forget set to remove specific knowledge. The update rule is as follows:
\[
\theta_{t+1} = \theta_t - \eta \left( \alpha \nabla_{\theta} L_{retain}(\theta_t) - \nabla_{\theta} L_{forget}(\theta_t) \right),
\]
where \( \theta_t \) denotes the model parameters at time step \( t \), \( \theta_{t+1} \) denotes the updated model parameters, \( \eta \) denotes the learning rate or step size, and \( \alpha \) is the retention coefficient that controls the influence of the retention data. \( L_{retain}(\theta_t) \) is the loss function evaluated on the retention set, and \( \nabla_{\theta} L_{retain}(\theta_t) \) is its gradient with respect to \( \theta_t \). \( L_{forget}(\theta_t) \) is the loss function evaluated on the forgetting set, and \( \nabla_{\theta} L_{forget}(\theta_t) \) is its gradient with respect to \( \theta_t \). This update rule effectively performs a scaled gradient descent on the retention set and gradient ascent on the forgetting set simultaneously, allowing for a controlled balance between knowledge retention and forgetting.

\paragraph{Representation Perturbation  (RMU) Method}
The Representation Perturbation Method (RMU) aims to disturb the learned representations of the model in order to encourage the forgetting of certain associations. The loss function encourages minimal difference between the model's representations before and after applying perturbations to the parameters:
\[
\mathcal{L}_{RMU}(\theta) = \mathbb{E}_{x \sim D} \left[ \| f(x, \theta) - f(x, \theta + \delta) \|^2 \right],
\]
where \( \mathcal{L}_{RMU}(\theta) \) denotes the loss function specific to the Representation Perturbation Method, \( x \) denotes the input data, \( \theta \) denotes the model parameters, \( f(x, \theta) \) denotes the model’s output representation for input \( x \) and parameters \( \theta \), \( \delta \) denotes the perturbation applied to the model parameters to disturb the representation.

\paragraph{Reinforcing Incorrect Answers (RIA) for Unlearning}
The Reinforcing Incorrect Answers (RIA) method aims to make the model "unlearn" harmful knowledge by encouraging it to predict incorrect answers for questions related to that harmful knowledge. This is achieved by training the model to decrease the loss associated with incorrect options, effectively making the model more likely to choose them. The loss function for RIA is defined as:
\[
\mathcal{L}_{RIA}(\theta) = - \sum_{i} \log \left( p(\hat{y}_i \mid x_i, \theta) \right),
\]
where \( \mathcal{L}_{RIA}(\theta) \) is the RIA loss function, \( x_i \) is the input question related to harmful knowledge for sample \( i \), \( \hat{y}_i \) represents the incorrect answer options for that question, and \( p(\hat{y}_i \mid x_i, \theta) \) is the probability assigned by the model with parameters \( \theta \) to these incorrect answer options. By minimizing this loss, we encourage the model to increase the probability of selecting incorrect answers.

\paragraph{Negative Preference Optimization (NPO) Method}
The Negative Preference Optimization method aims to reduce the likelihood of the model predicting incorrect outputs by minimizing the log-probability of unwanted outputs. This technique is effective in unlearning biased associations:
\[
\min_{\theta} \mathbb{E}_{x \sim D} \left[ \log \left( 1 - p(y \mid x, \theta) \right) \right],
\]
where \( \theta \) denotes the model parameters, \( D \) denotes the dataset distribution over input \( x \) and output \( y \), \( p(y \mid x, \theta) \) denotes the predicted probability of output \( y \) given input \( x \) and model parameters \( \theta \).

\section{Appendix B}

\subsection{Visualization of main experimental data}\label{sec:main_result}

\begin{figure*}[ht!] 
  \centering
  \includegraphics[width=\textwidth]{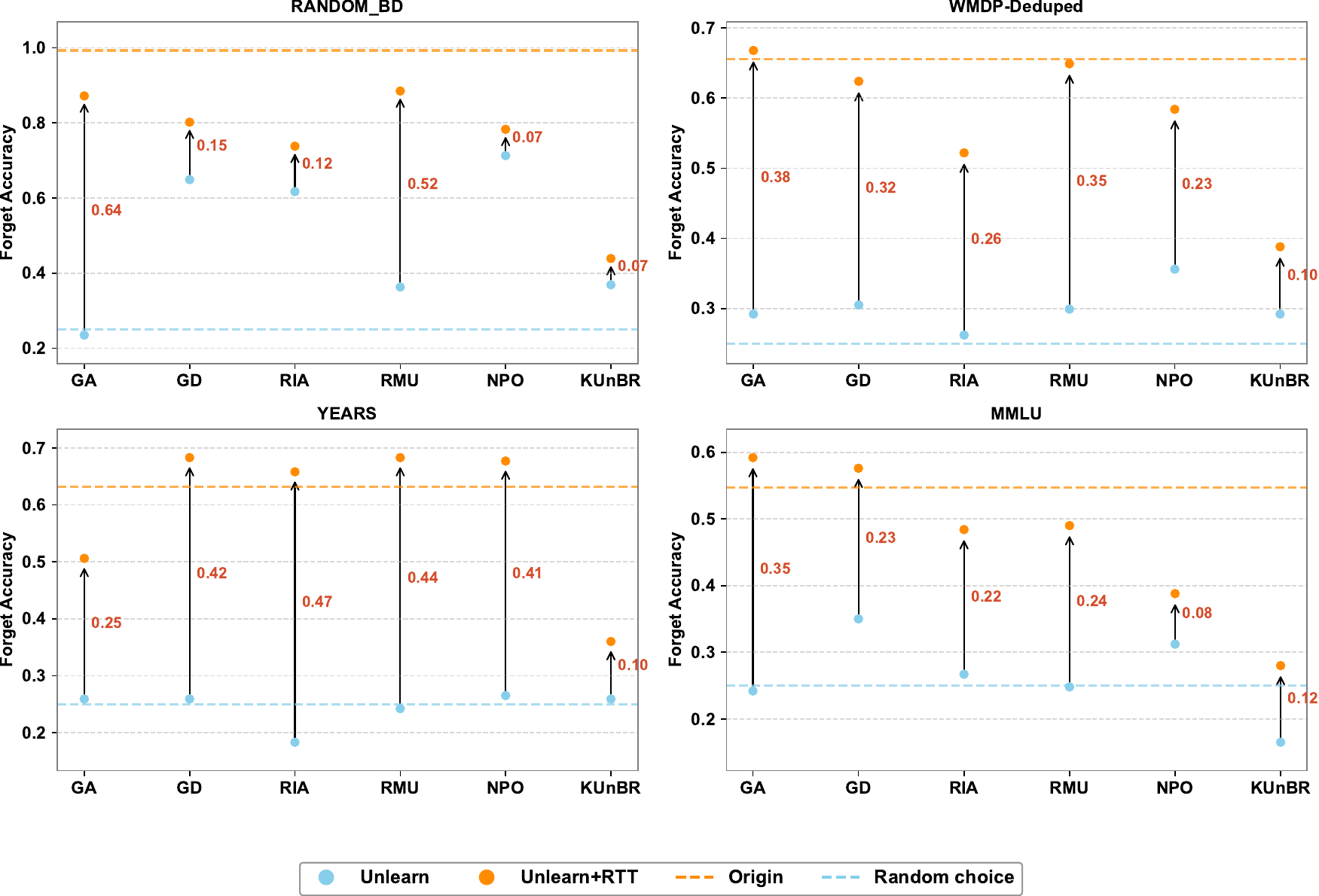}
  \caption{Comparison between our proposed \model and baselines when under RTT attack in terms of forget accuracy.}
\label{fig:main_result}
\end{figure*}

\begin{figure*}[ht!] 
  \centering
  \includegraphics[width=\textwidth]{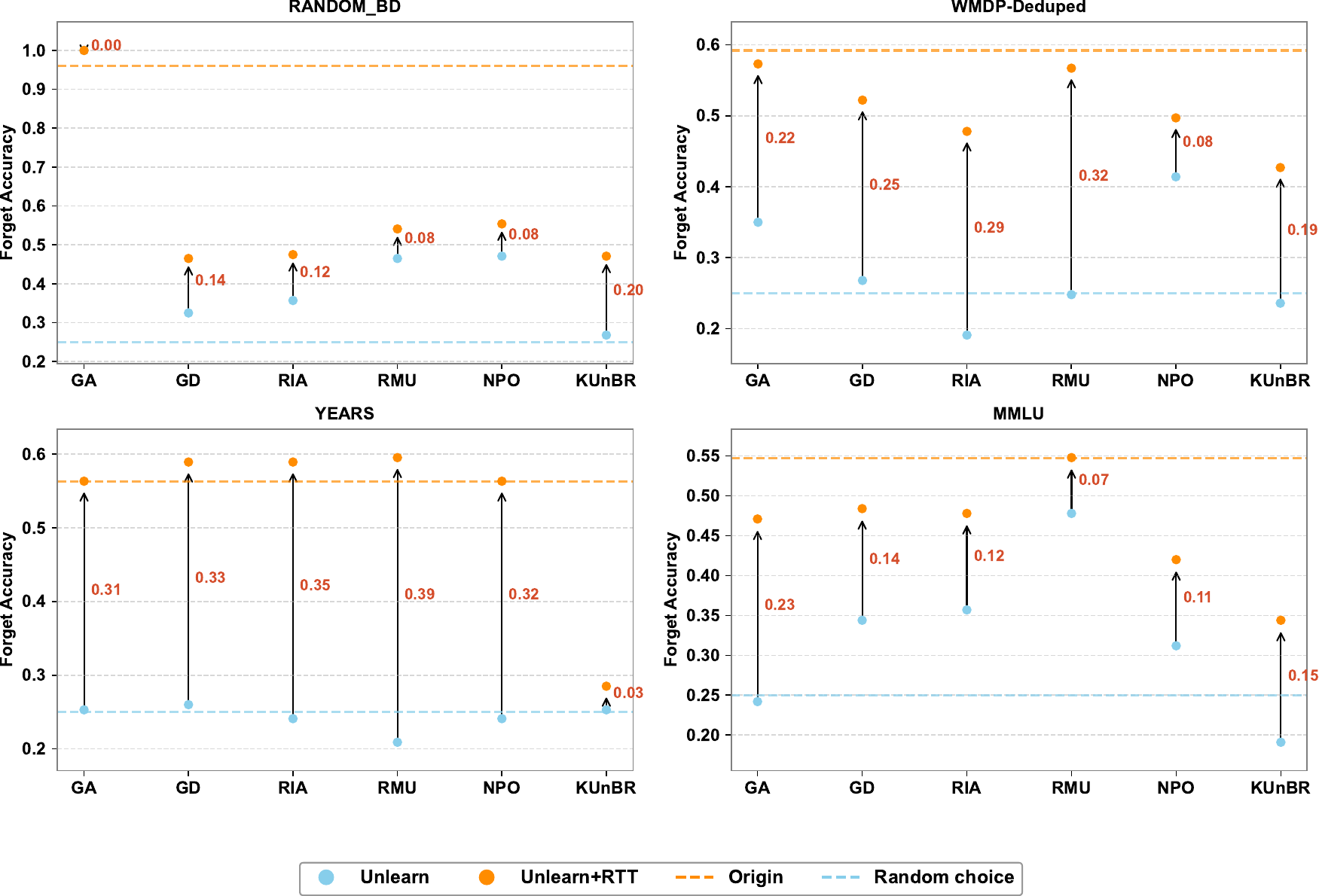}
  \caption{Comparison between our proposed \model and baselines when under RTT attack in terms of forget accuracy.}
\label{fig:main_result2}
\end{figure*}
To provide a more intuitive comparison, we include here a visual version of the main data from Figure~\ref{fig:main_result} and Figure~\ref{fig:main_result2}: the orange line shows the accuracy of the relevant data before unlearning, and the blue dashed line represents the random baseline. See Table~\ref{tab:main_result} for details of the figure.

\section{Appendix C}
\label{Appendix:Gradient Detail}
\subsection{Gradient Detail}
At present, some studies have demonstrated that unlearning can be achieved by fine-tuning only the parameters of the last few layers of the MLP. This observation suggests that the unlearning process might primarily influence the model’s output behavior, such as modifying its response to specific inputs without necessarily altering internal representations. However, our empirical results indicate that despite the high gradient magnitudes in the final layers, the actual unlearning performance remains suboptimal when restricting updates to the last two layers. This implies that effective unlearning may require intervention beyond surface-level output adjustments and necessitates modification of deeper or more distributed parameters. As a result, we exclude the last two layers from our block selection process.



\section{Appendix D}

\subsection{Experimental Hyperparameter Settings} \label{app:expsetting} 
To facilitate reproducibility and further research, we provide the source code of our method along with standardized implementations of all baseline approaches considered in this work. Each method is encapsulated into modular functions, allowing users to directly invoke unlearning procedures with minimal configuration. Our codebase also includes scripts for data preprocessing, model training, and evaluation under both standard and RTT settings.

\noindent The hyperparameters for \model are as follows: the learning rate (lr) is set to \(1.5 \times 10^{-7}\), the retention coefficient (retain coeff) is 0.1, and the warm-up step (warm step) is 24. Additionally, KUnBR uses a block number (block\_num) of M=8 and a block choice (block choose) of Top-K = 6 in 8 blocks.

\noindent For the other unlearning methods, the following hyperparameters are used: 
For GA, the learning rate is \(2.5 \times 10^{-7}\), the retention coefficient is 1, and the warm-up step is 24. For GD, the learning rate is \(1.5 \times 10^{-7}\), the retention coefficient is 1, and the warm-up step is 24. For RMU, the learning rate is \(1 \times 10^{-6}\), the retention coefficient is 10, and the warm-up step is 24. For RIA, the learning rate is \(2.5 \times 10^{-7}\), the retention coefficient is 2, and the warm-up step is 24. For NPO, the learning rate is \(8 \times 10^{-7}\), the retention coefficient is not specified (denoted by "-"), and the warm-up step is 24.

\end{document}